\documentclass{article}

\usepackage[final]{nips_2018}

\usepackage[utf8]{inputenc} % allow utf-8 input
\usepackage[T1]{fontenc}    % use 8-bit T1 fonts
\usepackage{hyperref}       % hyperlinks
\usepackage{url}            % simple URL typesetting
\usepackage{booktabs}       % professional-quality tables
\usepackage{amsfonts}       % blackboard math symbols
\usepackage{nicefrac}       % compact symbols for 1/2, etc.
\usepackage{microtype}      % microtypography
\RequirePackage{epsfig}
\RequirePackage{amssymb}
\RequirePackage{graphicx}
\usepackage{subcaption}
\usepackage{amsmath}
\usepackage[toc,page]{appendix}
\usepackage{verbatim}
\newcommand{\rpm}{\sbox0{$1$}\sbox2{$\scriptstyle\pm$}
  \raise\dimexpr(\ht0-\ht2)/2\relax\box2 }

\title{Improving Hospital Mortality Prediction with Medical Named Entities and Multimodal Learning}

\author{
\bf{Mengqi Jin, Mohammad Taha Bahadori, Aaron Colak,} \\
\bf{Parminder Bhatia, Busra Celikkaya, Ram Bhakta, Selvan Senthivel, Mohammed Khalilia,} \\\bf{Daniel Navarro, Borui Zhang, Tiberiu Doman, Arun Ravi, Matthieu Liger, Taha Kass-hout} \\\\
Amazon.com Services Inc.\\
\texttt{\{maggjin, bahadorm, parmib, busrac, hitenram, ssenthiv, khallia,}\\\texttt{navarda, bzha, tddoman, ravarun, ligerm, tahak\}@amazon.com} \\
\texttt{aaron.r.colak@gmail.com}
}

\begin{document}

\maketitle

\begin{abstract}
\begin{comment}
%Predictive modeling in health care area have been largely propelled by deep learning techniques in the past several years. 
%Studies in predicting clinical outcomes on benchmark tasks have shown promising results. 
Clinical text provides essential information to estimate the acuity of a patient during a hospital stay in addition to structured clinical data.
%Several studies showed that leveraging free text clinical notes can improve risk prediction performance. 
This study is a continuous work of a published benchmark using data derived from Medical Information Mart for Intensive Care (MIMIC-III). We mainly explore how clinical text can complement predictive learning. We preprocess clinical notes generated in the first two days of admission with an internal medical natural language processing service to extract entities of interest. A Doc2Vec algorithm is applied to train aggregated entities into vector representations. Further, we propose a multimodal network to train time series signals and unstructured clinical text jointly to predict the in-hospital mortality for ICU patients. By incorporating relevant medical entities with a multimodal network, our model outperforms the benchmark significantly for in-hospital mortality prediction task.
\end{comment}

Clinical text provides essential information to estimate the acuity of a patient during hospital stays in addition to structured clinical data. In this study, we explore how clinical text can complement a clinical predictive learning task. % by using data derived from Medical Information Mart for Intensive care (MIMIC-III). 
We leverage an internal medical natural language processing service to perform named entity extraction and negation detection on clinical notes and compose selected entities into a new text corpus to train document representations. We then propose a multimodal neural network to jointly train time series signals and unstructured clinical text representations to predict the in-hospital mortality risk for ICU patients. 
%We first process MIMIC-III notes generated in the first two days of admission with a medical natural language processing service to extract entities of interest. 
%Then, we apply a Doc2VecC algorithm to train aggregated entities into vector representations. 
%By incorporating medical entities with multimodal network, 
Our model outperforms the benchmark by 2\% AUC.

\end{abstract}

\section{Introduction}
% Paragraph I: thriving research in healthcare with deep learning
%Deep neural networks have transformed several application areas such as speech recognition \citep{2012Speech}, machine translation \citep{cite-machinetranslation}, and image caption generation \citep{imageDescription}. 
%As part of this revolution, 
A growing number of studies in the healthcare domain have shown compelling results by applying deep neural networks on predictive modeling tasks where traditional methods have met bottlenecks. 
Recurrent Neural Networks (RNN) and their variants such as Long Short-Term Memory (LSTM) were some of the earliest deep neural networks to analyze real-valued measurements \citep{LiptonKEW15, che2018recurrent} and higher dimensional structured claims data \citep{choi2016doctor}. Convolutional architectures also have been found to be faster and achieve similar accuracy \citep{razavian2016multi, liu2018deep}. In both cases, the learning objective was to stratify patients based on their risks of encountering certain clinical events, such as mortality and disease onset. 
Recently there has been increasing interest in utilizing unstructured information from clinical notes to improve clinical events prediction performance, as free-text notes paint a more elaborate picture of the patient.
% Patient representation & Clinical outcomes prediction
%Patient representation is one of the tasks that could be complemented by including text information. 
Topic modeling is commonly used to extract insights from notes \citep{Ghassemi2015AMT, miotto2016deep, rumshisky2016predicting, Suresh2017ClinicalIP}.  %\citep{CaballeroBarajas2015} 
%or sequence word embedding \citep{rajkomar2018scalable} before combining with other signals for learning. 
\citet{boag2018s} compared notes' representations generated by Bag of Words (BoW), Word2Vec and LSTM by evaluating their performance on downstream clinical prediction tasks. Their results showed no simple winning algorithm could ensure the best performance across all tasks. BoW perfomed well on tasks where outcomes were strongly correlated with certain phrases or words; LSTM, on the other hand, performed better on tasks for which temporal information was important. For in-hospital mortality prediction, BoW and Word2Vec achieved similar performance and outperformed LSTM. In this study, we choose a Doc2VecC algorithm \citep{cite-doc2vecc} to represent notes because of its proven success in capturing semantic information \citep{lau2016empirical} (details on text representation are in Section \ref{textrepresentation}).
% Other applicable area
%Apart from clinical event prediction applications, \citet{baumel2017multi} also tried to tackle diagnosis code assignment task by mining clinical narrative information, and achieved a state-of-art result with Hierarchical Attention-bidirectional Gated Recurrent Unit architecture.

%Whereas to our best knowledge, no studies had focused on the text preprocessing steps or leveraged different model architectures to model clinical narratives jointly with structured data. 

%\citep{miotto2016deep} showed how to learn a patient representation using stacked denoising autoencoder (SDA) \citep{vincent2010stacked} which further used to predict disease onset. 

%incorporating notes' information with topic modeling (\citet{Ghassemi2014}, \citet{Suresh2017ClinicalIP}), 
%ICU patients can generate rich data during their stays, including laboratory measurements, bedside observations, medications, charts and clinical notes, etc..

% Paragraph III: Difficult to share data & Experiment Repeatability 
%Meanwhile, since patients' health information is highly confidential, sharing patients' medical data is heavily regulated. 
%which brings obstacles for data sharing across organizations. 
%Patients' privacy are well protected under HIPPA. De-identification is required to remove PHI before any data sharing. 
%Moreover, various electronic health records (EHR) system vendors, different data storage models and customized implementation all contribute to preventing organizations from standardized data integration.
Meanwhile, results from using proprietary healthcare data are not easily reproducible, resulting in difficulties in making comparisons between different studies and algorithms. 
%The appearance of a publicly accessible critical care database Medical Information  Mart  for  Intensive  Care (MIMIC) \citep{cite-mimic} largely improved this environment. Whereas 
Outside of making data public and accessible, additional efforts are expected to ensure repeatable experiments, such as the well described study design and publicly available cohort construction code \citep{Johnson2017ReproducibilityIC}. Sharing the same goals, we selected a published in-hospital mortality prediction benchmark \citep{cite-benchmark} which uses only structured data (explained in Section \ref{learning}). We adopt the same cohort construction pipeline, and continuously iterate upon their model to ensure comparable and reproducible results.

% Paragraph V: Our focuses
Our work is distinguishable from other studies in three aspects:
1) We leverage a medical natural language processing service for named entity recognition and negation detection to process text beyond traditional text wrangling techniques; 2) We experiment with a multimodal model architecture to incorporate notes' information; 3) We also quantify the model's performance gain realized from the clinical notes (which are directly comparable to the published benchmark).
%1) how much can clinical texts complement structured data on in-hospital mortality prediction; 2) what is a better way to utilize text information; 3) will using embeddings trained on named entities make a difference on performance comparing to those trained on simple tokenized text. 

%\begin{enumerate}
%    \item We incorporated clinical text information in the model to explore how much potential value could be introduced by clinical notes.
%    \item We compared two model architectures to leverage a better way to utilize unstructured clinical text on mortality prediction task.
%    \item We compared two text preprocessing approaches to test their impact on the benchmark task.
%    \item We continuously worked on a published benchmark to generate comparable results.
%\end{enumerate}

%%%%%%%%%%%%%%%%%%%%%%%%%%%%%
% Background and Related Work
%%%%%%%%%%%%%%%%%%%%%%%%%%%%%
% For now
% \section{Background and related work}
%\input{related.tex}

%%%%%%%%%%%%%%%%%%%%%%%%%%%%%
% Cohort
%%%%%%%%%%%%%%%%%%%%%%%%%%%%%
\section{Data extraction and text representation}
\begin{figure}[t]
  \centering 
  \includegraphics[width=5.5in]{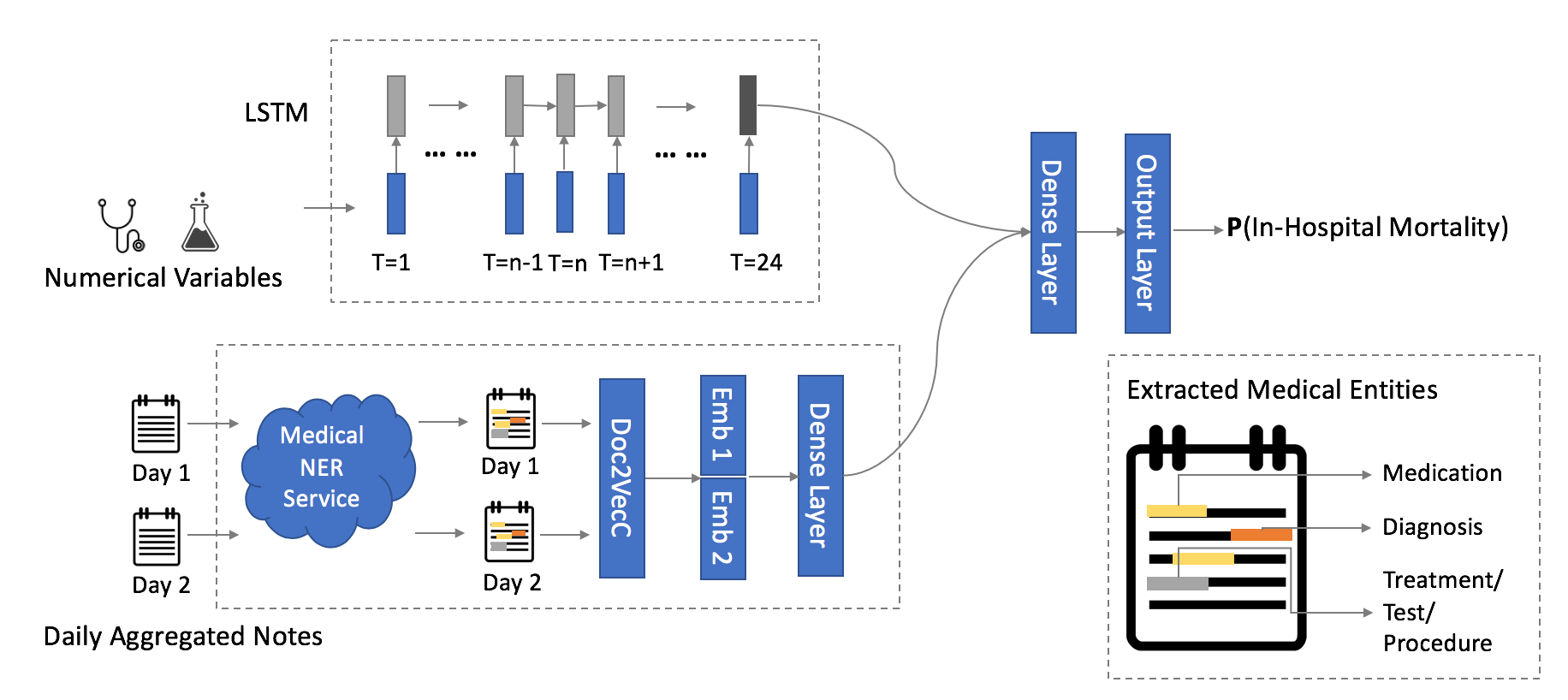}
  \caption{Data processing and multimodal architecture of proposed model: selected named entities are extracted from daily aggregated notes to train document embeddings to pass through a shallow feedforward neural network before jointly trained with last output from LSTM to make predictions.}
  \label{fig:pipeline}
\end{figure}

We conduct experiments on MIMIC-III data set \citep{cite-mimic}. 
%The data %is collected at Beth Israel Deaconess Medical Center, 
%contains 58,000+ ICU stays of 46,000+ patients. % from 2001 to 2012. 
%Both structured data and clinical notes are used to train the model. 
%\subsection{Cohort construction criteria} 
We adopt the same data extraction pipeline as the benchmark \citep{cite-benchmark} to prepare baseline features and introduce note assignment events. The data analysis flowchart of the best-performed model is pictured in Figure \ref{fig:pipeline}.  Overall, there are 42,276 ICU stays extracted from 33,798 patients at least 18 years old at admission. 
First 48 hrs data of ICU admission are collected to predict whether the patient encounters a fatality throughout her stay. 
The same split of 70\%/15\%/15\% and seed state are used to create train/validation/test sets. 

\subsection{Baseline feature set}
The benchmark cohort contains 17 signals derived from the structured database including vital signs and lab results. In the fixed 48 hrs observation window, signals are discretized every 2 hours, which yields 24 signals per stay. Missing values are imputed from the previous record, and population statistics if the previous record was unavailable. This feature set is referred to as \textbf{Vital}.

%\begin{table}[htbp]
%  \caption{Cohort Summary Statistics}
%  \label{tab:cohort}
%  \centering 
%  \begin{tabular}{llll}
%  \toprule
%               & Cases           & Controls & Total \\ \midrule
%    Training   & 1,987 (13.47\%) & 12,760   & 14,747 \\
%    Validation & 436 (13.46\%)   & 2,804    & 3,240 \\  
%    Test       & 374 (11.52\%)   & 2,873    & 3,247 \\
%    Total      & 2,797 (13.17\%) & 18,437   & 21,234 \\
%  \bottomrule
%  \end{tabular}
%\end{table}

\subsection{Clinical text representation} \label{textrepresentation}
We first examine the completeness of timestamps associated with the note to propose model structures. ECG and Echo reports do not have time stamps, but only a date available, which take up 10.49\% of the first two days' non-discharge notes. In light of the incompleteness of date-only time-stamped notes, we concatenated clinical notes assigned in the first two days of admission into two aggregated notes. Discharge notes are excluded from the text corpus to avoid information leak. We replaced numbers in notes with zeros and low-frequency words that  have appeared less than 10 times with a special token.
%A selected Doc2VecC algorithm (\citet{cite-doc2vecc}) was applied to train text into vector representations. 
A Continuous Bag-Of-Words based document representation algorithm, Document Vector through Corruption (Doc2VecC) \citep{cite-doc2vecc} is selected to learn note embeddings.
%Doc2VecC algorithm represents each note by averaging over its word embeddings to reflect semantic meanings. 
In this approach, the original document is corrupted by randomly removing a significant portion of words, and the document is represented using only the embeddings of the remaining words. We select Doc2VecC because it offers speedup during training as it significantly reduces the number of parameters to update in backpropagation. Another advantage of this method is that weighing down frequent words across document corpora can yield performance improvements. 

\vspace{-9pt}
\paragraph{Embeddings of tokenized notes}
We use an internal medical text tokenization service to process daily aggregated notes. We then represent tokenized notes using Doc2VecC algorithm after following the preprocessing steps introduced above and obtain a vocabulary size of 108,907. This set of notes' representations is referred to as \textbf{NoteEmb}. 
% note
% Vocab size: 251,045
% Words in train file: 389053731
% note 2
% Vocab size: 108,907
% Words in train file: 640523613
% entity
% Vocab size: 25,409
% Words in train file: 53141648
% entity 2
% Vocab size: 25,503
% Words in train file: 53141648

% An example to write note processing details
%We represent patients with a concatenation of all their non-discharge notes. We tokenize the data using the Ucto tokenizer (Van Gompel et al., 2012) and lowercase it. To obtain patient representations using the SDAE, we replace the numbers, and certain time and measurement mentions with special tokens.
%We remove the punctuations, and the terms with frequency < 5. We use a bag-of-words (BoW) with their TF-IDF scores as features, to obtain a vocabulary size of 71,001. We also conducted the experiments with a bag-of-medical-concepts feature set, but they performed consistently worse. To train the doc2vec models, we remove the numbers, and the tokens matching time and measurement patterns (determined from the initial validation set results), and get a vocabulary size of 48,950.

\vspace{-9pt}
\paragraph{Embeddings of concatenated relevant medical entities}
% Importance to use NER and negation to process notes

%BC: Clinical text could be full of noisy and redundant information thus requires proper prepossessing, such as named entity recognition, normalization, and negation tagging \citep{miotto2015case}. 
%BC: We experiment with training note embeddings on top of text body composed of medical relevant entities to achieve competitive performance on mortality prediction task. 
Clinical text can be full of redundant information, therefore requires proper pre-processing such as named entity recognition (NER), and negation scope detection to filter negated information \citep{miotto2015case}. 
Contrary to previous studies \citep{miotto2016deep, liu2018deep} that tag clinical narratives using medical ontology look-up and regular expressions to detect negations, such as NegEx \citep{chapman2001simple}, in this paper, we leverage a neural network based internal medical NER service \citep{amazonNER}. The NER system has a hierarchical architecture composed of three components: 1) a convolutional character-level encoder extracting features for each word from characters, 2) a convolutional word-level encoder extracting features from the surrounding sequence of words, and 3) an LSTM tag decoder inducing a probability distribution over any sequences of tags. 
We extract five types of entities including medical condition, medication, tests, treatments, and procedures. The medical NER service also jointly identifies negated entities and returns a negation tag as an attribute. We then exclude negated entities from the text corpus before training embeddings. %, as absent medical condition is not clinically relevant \citep{plaza2010retrieval}.
% Giving an example
For example, a phrase "no oropharyngeal lesion" has "oropharyngeal lesion" recognized as a medical condition type of entity with negation attribute. This entity is discarded from the text corpus as it indicates an absence of the condition.
The vocabulary size of extracted entities corpus is 25,503. Embeddings trained on top of concatenated named entities is referred to as \textbf{EntityEmb}.
% Amazon NER
\vspace{-0.1in}
%BC: The internal NER system is designed for efficiently active learning with a new network architecture composed of three components: 1) a convolutional character-level encoder extracting features for each word from characters, 2) a convolutional word-level encoder extracting features from the surrounding sequence of words, and 3) an LSTM tag decoder inducing a probability distribution over any sequences of tags. 

%%%%%%%%%%%%%%
% Methods
%%%%%%%%%%%%%%
\section{Learning}
We experiment with two neural network architectures to model the medical text. We first apply a LSTM on concatenated baseline features with pre-trained text embeddings. 
%During time interval when the note with max note ID is set to the aggregated note/entity embedding vector. 
Daily concatenated notes are assigned timestamps until the end of the day. Missing embeddings are imputed in the same way as baseline features. 
Second, we test a multi-modal network to process structured signals and embeddings separately, where baseline features are passed through LSTM, and two pre-trained daily embedding vectors are concatenated to pass through a feedforward network and are jointly trained afterward.
\vspace{-9pt}
\paragraph{Benchmark model architecture} 
%Recurrent Neural Network (RNN) is a type of deep learning 
The benchmark model is a LSTM neural network and uses only structured data. 
%The benchmark uses LSTM neural network \citep{cite-lstm} with peephole connections \citep{peepholeLSTM} to learn baseline features. 
%LSTM is one type of RNN and famously known for its capabilities to capture both long-term and short-term dependencies in sequence data (\citet{cite-lstm}). 
Given a series of evenly spaced signals within a fixed observation window, $\mathbf{x}=\{\mathbf{x}_1, \mathbf{x}_2, ..., \mathbf{x}_T\}$, where $\mathbf{x}_t \in \mathbf{R}^L$ and $L$ is the number of input signals, the model learns a series of hidden state vectors $\mathbf{h}_1, ..., \mathbf{h}_T = \mathrm{LSTM}(\mathbf{x}_1, ..., \mathbf{x}_T)$, and uses the last hidden vector $\mathbf{h}_T$ to predict outcome label $y$.

\vspace{-9pt}
\paragraph{Proposed multimodal neural network}
%In order to compare performance cross different feature set, algorithms and model structures, it is important to compare models and methods on same data sets to arrive at a right conclusion. For this reason, we adopted the same data processing steps as \citet{cite-benchmark} for all the experiments carried on in this paper. Han
Multimodal deep learning frameworks were introduced by \citep{cite-multimodal}. When multiple modalities are available, the fused representation can be used as input for discriminative tasks \citep{multimodal-DBM}. 
We concatenate two days' text embeddings $\mathbf{e}_1$ and $\mathbf{e}_2$, and pass it through one layer feedforward network to learn a hidden representation $\mathbf{h}_e$. Next, we concatenate $\mathbf{h}_e$ and last output from LSTM $\mathbf{h}_T$ to learn a joint representation $\mathbf{h}_j$ of clinical variables and text embeddings through a feedforward network to predict class label $y \in {0, 1}$.

%\begin{align*}
%    \mathbf{h}_e &= \mathbf{W}_e\cdot[\mathbf{e}_1, \mathbf{e}_2] + b_e\\
%    \mathbf{h}_j &= \mathbf{W}_j\cdot[\mathbf{h}_T, \mathbf{h}_e] + b_j\\
%    \hat{y} &= \sigma(\mathbf{W}_y \cdot \mathbf{h}_j + b_y)
%\end{align*}

%The loss function :
%\begin{align*}
%    CE(y,\tilde{y}) = -(y \cdot log(\tilde{y})+(1-y) \cdot log(1-\tilde{y}))
%\end{align*}
%Include the diagram of the model architecture.
 \label{learning}

%%%%%%%%%%%%%%
% Results
%%%%%%%%%%%%%%
\section{Experiments} 
%Experiments results show that our proposed model combining named entities embeddings achieves the highest performance comparing to the benchmark study set up locally and other studies on mortality prediction task. 

\subsection{Experimental Setup}
We establish the selected benchmark with the same parameter settings (\textbf{Vital}) \citep{cite-benchmark}. 
%Then we train models across two combination of features set (\textbf{Vital + NoteEmb} and \textbf{Vital + EntityEmb}) on two different neural network archtectures: LSTM and Multimodal architecture. 
We then compare mortality prediction performance of our two neural network structures (LSTM and Multimodal) using two feature sets (\textbf{Vital + NoteEmb} and \textbf{Vital + EntityEmb}).
%We tune hyperparameters including hidden layer size of LSTM, hidden layer size of feedforward network, learning rate, dropout and L2 regularization. 
No dropout nor normalization is applied. The hidden layer size of LSTM is set to 256. The number of hidden units of the dense layers for text representation and joint representation are set to 100 and 300 respectively. Parameters are cross-validated on the validation set. Learning rate is set to 0.0001. 
%We use Adam optimization \citep{AdamKingmaB14} on mini-batches of 8 examples. 
%The model with lowest validation loss is selected as the best model among those generated with the same initialization. 
%Cohort construction code is available at (Amazon Github). 
Our experiments are implemented with Lasagne version 0.2.1 \citep{cite-lasagne}, and run on Amazon Web Services' p2. 8xlarge GPU instances. 

%\subsection{Evaluation}
%Two metrics used to evaluate model performance are area under the receiver operator's curve (AU-PRC) and area under precision-recall curve. As, the overall mortality rate on the data set is 11\%, given this imbalanced outcome distributions, AU-PRC is known to be robust on imbalance data set (\citet{cite-aurocprc}). 
%For each model, we train the model 20 times with different initialization seeds. The best performed model on the validation set was selected by F1 score, and bootstrapped on the test set with replacement over 100 draws.

\subsection{t-SNE plot of medical text embeddings} 
\begin{figure}[t]
\vspace{-0.2in}
\centering
\begin{subfigure}[t]{.5\textwidth}
  \centering
  \includegraphics[width=2in]{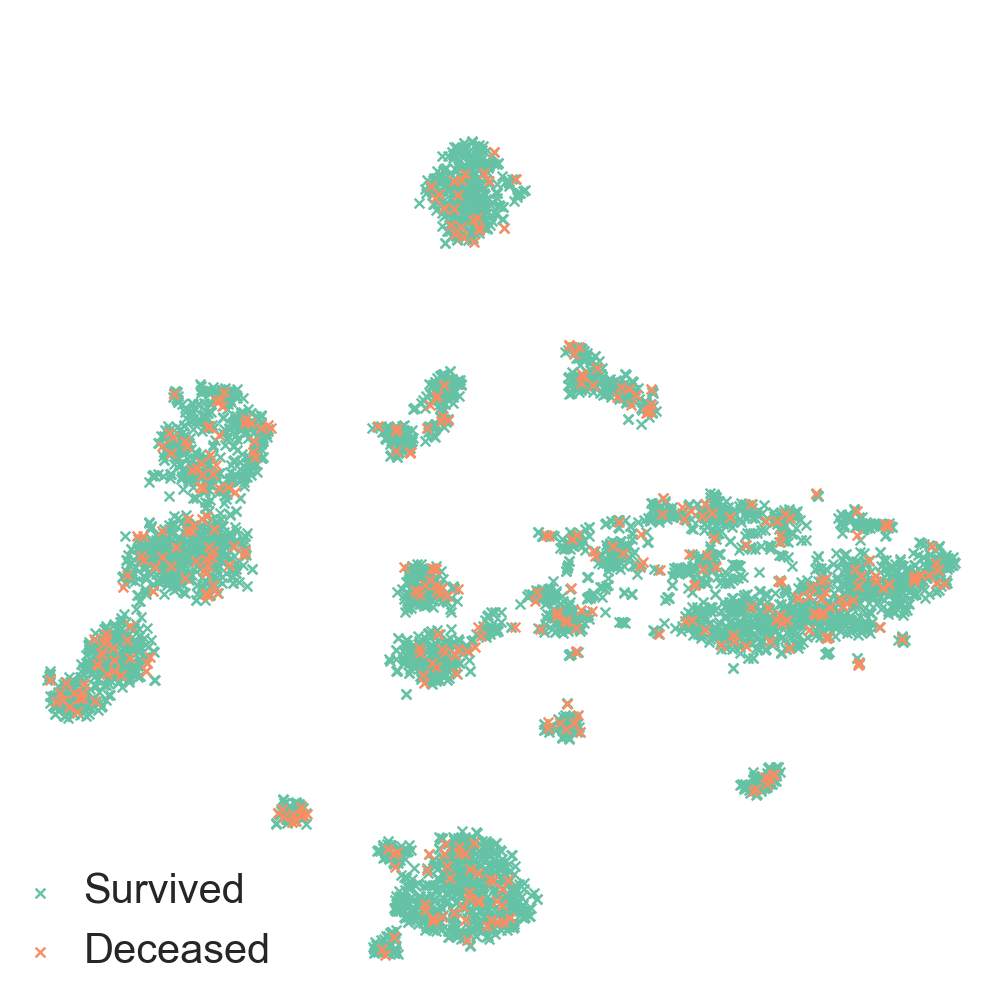} 
  \caption{aggregated raw notes embeddings.}
  \label{fig:tsne-note} 
\end{subfigure}%
~
\begin{subfigure}[t]{.5\textwidth}
  \centering
  \includegraphics[width=2in]{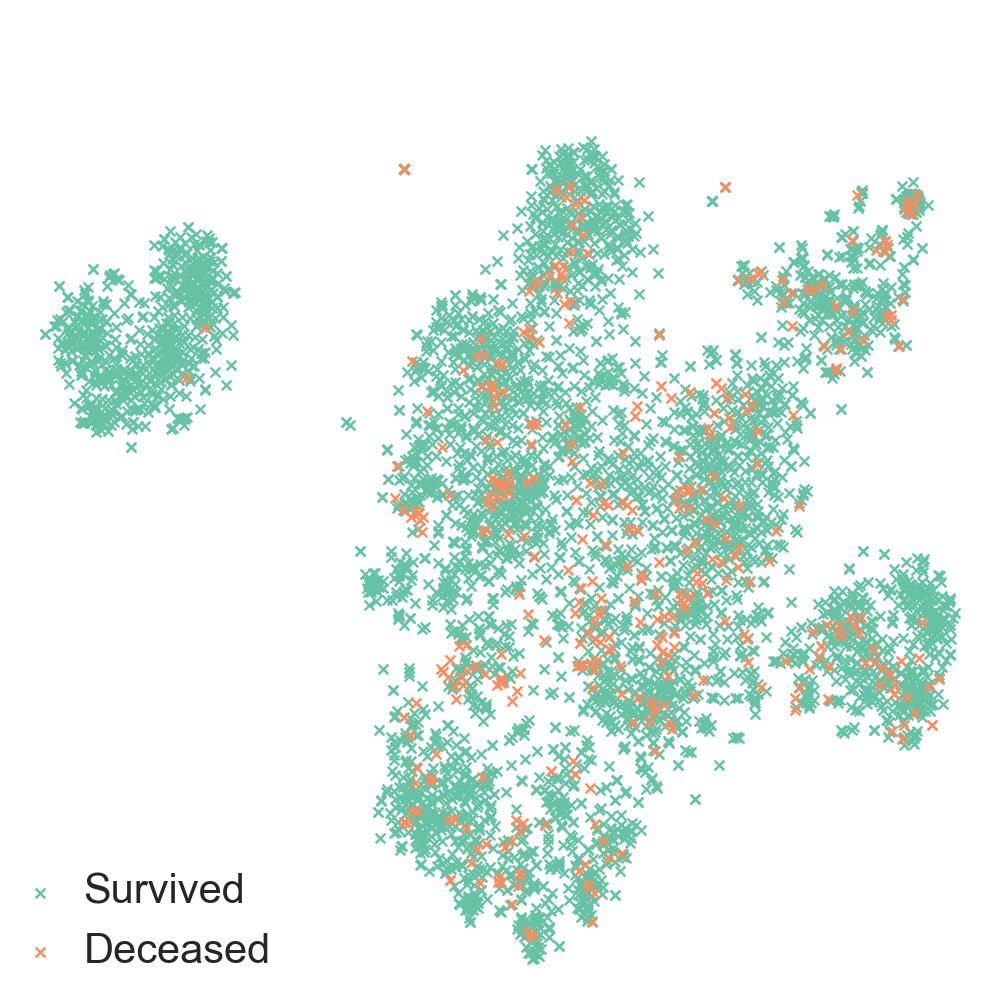} 
  \caption{aggregated entity embeddings.}
  \label{fig:tsne-ner}
\end{subfigure}
\caption{t-SNE plots for embeddings, color-coded by associated outcomes.}
\label{fig:tsne}
\vspace{-0.1in}
\end{figure}

We apply the t-SNE algorithm \citep{cite-tsne} to visualize embeddings generated by the Doc2VecC algorithm. 
We show two t-SNE plots for notes' embeddings (Figure \ref{fig:tsne-note}) and aggregated named entities' embeddings (Figure \ref{fig:tsne-ner}). 
Embedding vectors are colored based on their associated clinical outcomes. 
It’s interesting to see that the notes embeddings and entities embeddings are projected into different patterns. 
Moreover, a cluster of entities embeddings is mostly associated with negative outcomes. Whereas, the positive and negative cases can be observed in most subgroups of notes embeddings.
This suggests embeddings generated from entities correlates with the final outcome of an ICU admission.

\subsection{Prediction results} 
We train each model 20 times with different initialization seeds. 
The best-performed model on the validation set is selected by the F1 score, and bootstrapped on the test set over 100 re-sampling. 
%Results show that incorporating notes information with entity embedding helps to improve the performance given the same network structure; multimodal learning boosts the performance on the same feature set (Table \ref{tab:results}). 
%Overall, combining aggregated named entities embeddings with structured data trained by multimodal network yields the highest performance (AU-ROC 0.8730 [95\% CI 0.8715-0.8745] and AU-PRC 0.5451 [95\% CI 0.5502]), increasing in-hospital mortality performance  by 1.99\% in AU-ROC and 4.21\% in AU-PRC comparing to the benchmark (AU-ROC 0.8531 [95\% CI 0.8521-0.8551] and AUPRC 0.5030 [95\% CI 0.0.4979-0.5081]).
%We focus on how to use free text information to build competitive risk models. %not just to prove the potential of clinical notes, but also explored how to use them effectively.
Our results show the proposed multimodal neural network is more efficient to incorporate text information with structured data comparing to LSTM. Meanwhile, models using embeddings trained from medical named entities present consistently better performance than those trained from the simple tokenized text. Our results are directly comparable and reproducible by adopting the same data processing flow as the benchmark study. In the future, we would like to further apply entity linking techniques to normalize extracted entities, and explore different algorithms to represent notes information to enhance predictive modeling of clinical events. 
%On the other side, more accurate prediction on mortality in ICU can benefit healthcare providers to have a better control of the acuity and potential outcomes of patients in early stage of admission.
\begin{table}[t]
\caption{Experimental Results}
  \centering
  \begin{tabular}{lllll}
    \toprule
    \multicolumn{2}{c}{Model}                   \\
    \cmidrule(r){1-2}
    Feature Set                & Neural Network Structure   & AU-ROC (\%) & AU-PRC (\%) \\
    \midrule
    Vital (Benchmark)                    & LSTM                       & 0.8531 \rpm 0.0020   &  0.5030 \rpm 0.0051   \\
    Vital + NoteEmb            & LSTM                       & 0.8496 \rpm 0.0018  &  0.5040  \rpm 0.0050  \\
    Vital + NoteEmb            & Multi-modal                & 0.8669 \rpm 0.0018  &  0.5310  \rpm 0.0051  \\
    Vital + EntityEmb          & LSTM                       & 0.8703 \rpm 0.0017   &  0.5470 \rpm 0.0048  \\
    Vital + EntityEmb          & Multi-modal                & \bf{0.8734 \rpm 0.0019}   &  \bf{0.5290 \rpm 0.0056}    \\
    \bottomrule
  \end{tabular}
  \vspace{-0.1in}
  \label{tab:results}
\end{table}

%%%%%%%%%%%%%%
% Discussion
%%%%%%%%%%%%%%
%\section{Conclusions} 
%\input{discussion.tex}

%\subsubsection*{Acknowledgments}
%Use unnumbered third level headings for the acknowledgments. All acknowledgments go at the end of the paper. Do not include acknowledgments in the anatomized submission, only in the final paper.

\medskip

\small
\bibliography{references}

%\begin{appendices}
%\chapter{Non-discharge Notes Statistics}
%\input{appendix_1.tex}
%\end{appendices}
\end{document}